\documentclass[runningheads]{llncs}
\usepackage{graphicx}
\usepackage{listings}
\usepackage{subfig}
\usepackage{xcolor}
\usepackage{hyperref}
\usepackage{cleveref}
\usepackage{tikz}

\begin{document}
\title{SOTER on ROS: A Run-Time Assurance Framework on the Robot Operating System}
\titlerunning{SOTER on ROS}
%
\author{Sumukh Shivakumar\inst{1}\and Hazem Torfah\inst{1}
\and Ankush Desai\inst{2}\and
Sanjit A. Seshia\inst{1}}
\authorrunning{S. Shivakumar et al.}
%
\institute{University of California, Berkeley, USA \\
\email{\{sumukhshiv,torfah,sseshia\}@berkeley.edu}\\
 \and
Amazon, USA\\
\email{ankushpd@amazon.com}}
\maketitle              
\begin{abstract}
We present an implementation of SOTER, a run-time assurance framework for building safe distributed mobile robotic (DMR) systems, on top of the Robot Operating System (ROS). The safety of DMR systems cannot always be guaranteed at design time, especially when complex, off-the-shelf components are used that cannot be verified easily.  SOTER addresses this by providing a language-based approach for run-time assurance for DMR systems. SOTER implements the reactive robotic software using the language P, a domain-specific language designed for implementing asynchronous event-driven systems, along with an integrated run-time assurance system that allows programmers to use unfortified components but still provide safety guarantees. We describe an implementation of SOTER for ROS and demonstrate its efficacy using a multi-robot surveillance case study, with multiple run-time assurance modules. Through rigorous simulation, we show that SOTER enabled systems ensure safety, even when using unknown and untrusted components.
\end{abstract}
\section{Introduction}
The design of runtime monitoring components has become an integral part of the development process of distributed mobile robotic (DMR) systems. Runtime monitoring is essential for maintaining situational awareness, assessing the health of a robot, and most importantly for detecting any irregularities at runtime and consequently deploying the necessary countermeasures when such irregularities occur. 
The growing complexity of DMR systems, along with the utilization of uncertified off-the-shelf components and complex machine-learning models that are difficult to verify at design time, has made runtime assurance a crucial component for building robust DMR systems~\cite{seshia-arxiv16}.

\begin{figure}[t]
  \centering
  \includegraphics[scale=0.26]{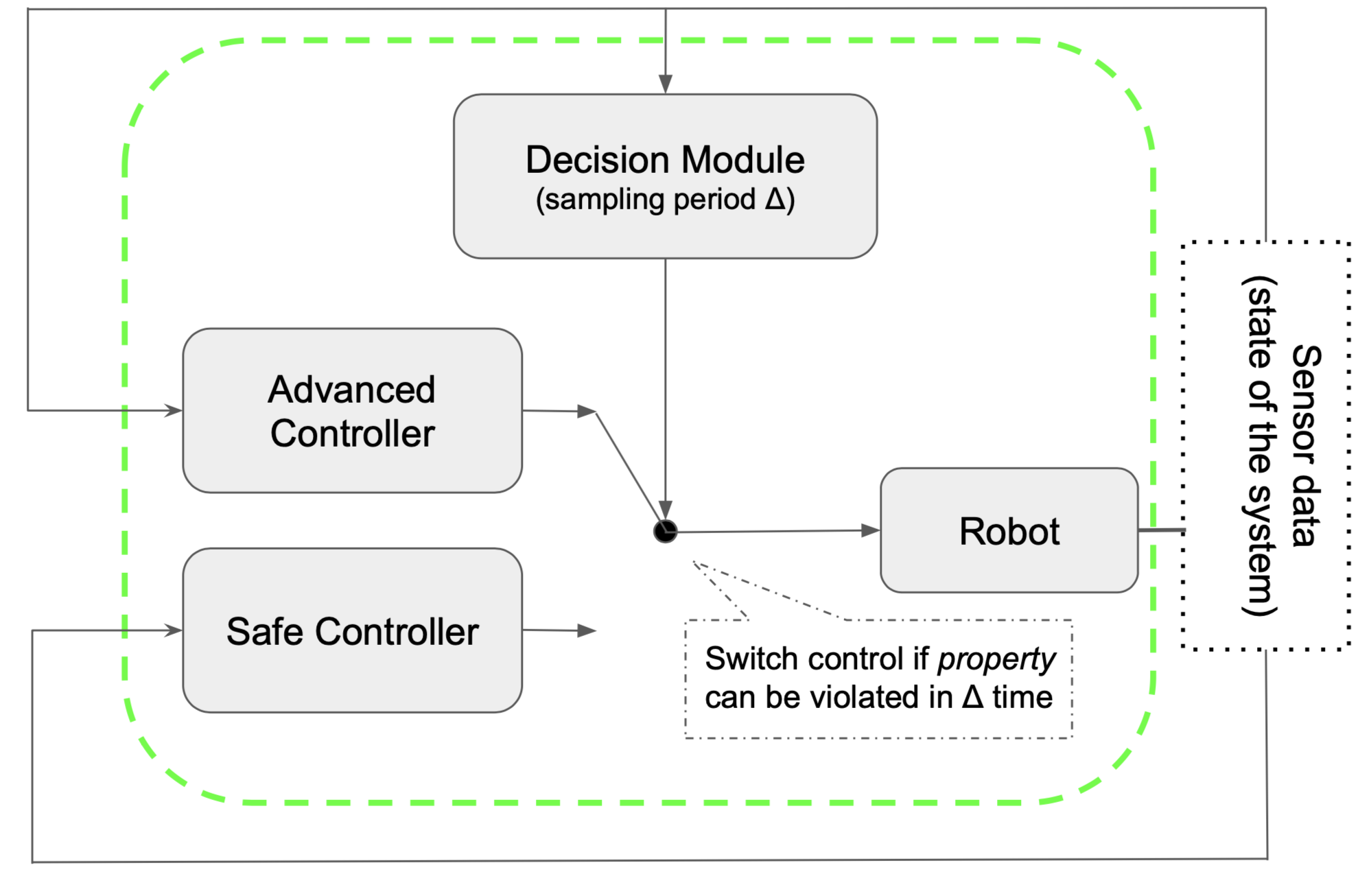}
  \caption{RTA module in SOTER.}
    \label{fig2: Simplex Architecture}
\end{figure}

In this paper, we present an implementation of SOTER~\cite{soter}, a runtime assurance framework for building safe distributed mobile robotics, on top of the Robot Operating System (ROS)\footnote{\url{https://www.ros.org}}. 
In SOTER, components of a DMR system are defined as runtime assurance (RTA) modules implementing a Simplex architecture \cite{simplex}.
An RTA module based on Simplex (see \Cref{fig2: Simplex Architecture}) consists of two controllers, an \emph{advanced controller} (AC) and a \emph{safe controller} (SC), and a \emph{decision module} that implements a switching logic between the AC and SC. The AC is used for operating the system under nominal circumstances. This is usually an optimized controller based on advanced heuristics or complex learning-enabled components such as machine-learning-based perception modules. This makes it hard to provide any guarantees on the behavior of the AC, especially, when it is an off-the-shelf component that cannot be verified at design time. To, nevertheless, guarantee the safety of a system using such controllers, the system can always default to a certified back-up controller, the SC, that takes over operating the system when anomalies in the behavior of the AC are detected. For example, the SC could be based only on reliable sensors that navigate a robot to a safe state. The detection of faulty behavior is guaranteed by the decision module, which is a certified monitor that observes the state of the robot. The decision module decides whether it is necessary to switch from the AC to the SC to keep the robot in a safe state and when to switch back to the AC to utilize the high performance of the AC to optimally achieve the objectives of the robot. In DMR systems, components within the robot as well as any systems connected to the robot are communicating asynchronously. In SOTER, the various robot components are implemented as asynchronously communicating RTA modules.  This is realized by implementing the modules in the language P \cite{P}, a verifiable programming language designed for writing asynchronous event-driven code, which can be compiled down to code executable on platforms such as widely used platforms as the Robot Operating System (ROS).

The implementation of SOTER presented in this paper maintains a similar approach to implementing the robot components as RTA modules with the following new extensions: 
\begin{itemize}
    \item A refactorization of SOTER to support portability onto various Robot SDK's. 
    The refactorization separates the software stack implementing the robot from the used robot SDK. Implemented in P, this allows us to provide a robot implementation with formal guarantees on the behavior of the interacting robot components. This also allows us to easily port this framework on to other robot SDK's.
    \item The refactorization also includes a separation between the implementation of robot's RTA modules' logic and the actual AC and SC implementations used in these RTA modules. 
    This allows us, in a plug-and-play fashion, to easily link the calls of the AC's and SC's in the RTA modules to external implementations of the controllers.  
    \item A concrete integration of the SOTER framework onto widely used robot SDK ROS. We provide an implementation of a software interface that implements the communication with the ROS SDK. Integration onto other robot SDK's can be done in a similar way. 
\end{itemize}

The implementation of the framework, details and videos on the examples presented in \Cref{sec:case_studies}, and a guideline for using the framework can be found on the following website \url{https://github.com/Drona-Org/SOTERonROS}. This includes instructions on how to execute the examples presented in the paper.

\section{Architecture of SOTER on ROS}
\label{sec:arch}
\begin{figure}[t]
  \centering
  \includegraphics[scale=0.27]{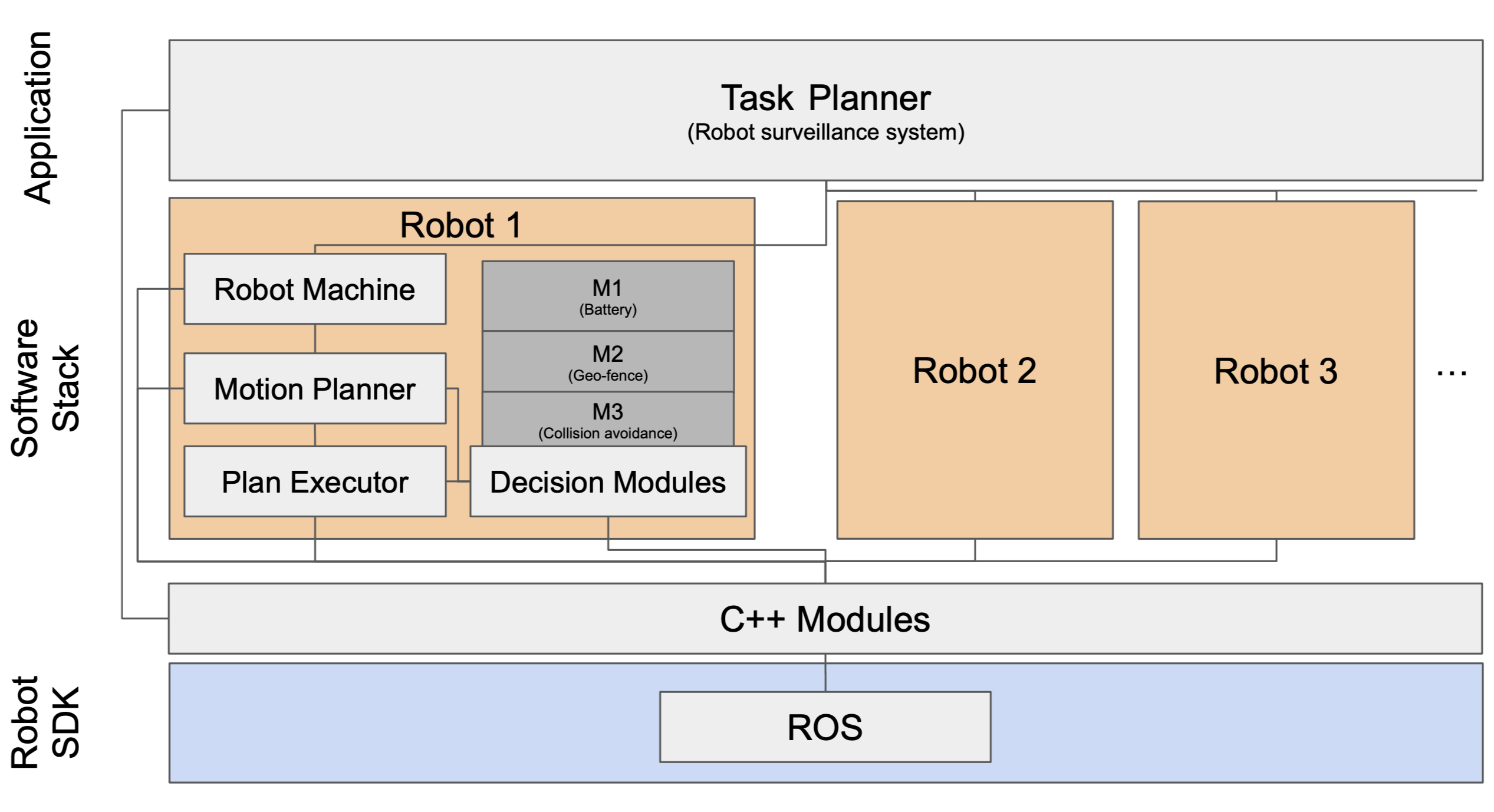}
  \caption{SOTER Framework Architecture.}
    \label{fig1: SOTER Architecture}
\end{figure}

In this section, we outline the architecture of the new implementation of the SOTER framework. The framework borrows from the Drona framework \cite{drona} and is similarly comprised of three layers. The \emph{application} layer, the robot \emph{software stack}, and the \emph{robot SDK}. The application layer implements an application related task planner that is responsible for computing application related tasks and distributing them amongst the robots for execution.  
The \emph{software stack} of each robot consists of an interface to establish the communication with task planner, a motion planner, and a plan executor, in addition to a set of decision modules. In contrast to the Drona framework, the motion planner and the plan executor are implemented as RTA modules that are linked to one of the decision modules and to implementations of their safe and advanced controllers. 

The implementation of the monitors used by the decision modules, and also the implementation of the safe and advanced controllers for both the motion planner and the plan executor are provided as C++ modules in a separate library. The library also plays the role of the interface which abstracts away many of the underlying details needed for the robot SDK, and make them accessible to the modules in the software stack as well as to the task planner.

In the following, we give some details on the implementation of each of the three layers and the integration into robot SDK's such as ROS.  We use a robot surveillance application to elaborate on some of the implementation details. 

\paragraph{Task Planner.}
A task planner is implemented specifically for a certain application. For example, a task planner for a surveillance application computes certain way-points on the map that should be visited by the robots. 
A task planner in our framework is implemented as a state machine in the language P \cite{P}. This allows asynchronous communication between the task planner and P state machines defining the robots.  A state machine in P has its own event queue on which it receives events published by other machines that communicate with this machine. The task planner can send events defining tasks to the queues of the robot machines for execution.   In its initial state, the task planner state machine,  spawns a number of robots to execute application related tasks. After initializing the robots, the  task planner computes the tasks to be executed and publishes the tasks to the event queues of the different robots.

\paragraph{Robot Software Stack.}
The software stack consists of three predefined P state machines, the robot machine, the motion planner, and the plan executor, in addition to other application-dependent P state machine defining the decision modules used by the motion planner and the plan executor. 
When the task planner spawns a robot, a new software stack is setup with a new robot machine, motion planner, plan executor and all the decision modules, 
In the following, we provide some details on the state machines defining the robot machine, motion planner, and plan executor. All implementations can be found on the frameworks webpage under the software stack directory. 

The robot machine serves as the interface between the task planner and the motion planner of a robot. The tasks assigned by the task planner are queued in the event queue of the robot machine.  When an event is received, the robot machines processes the event and forwards the event to the queue of the motion planner. 
%
For each task, processed and sent by the robot machine, the motion planner computes a plan to execute this task. For example, in the robot surveillance application, the tasks are destinations the need to be visited and the plan would be a series of way-points to reach each destination. The state machine processes the tasks one by one. For each task a plan is computed and then sent to the plan executor. A plan for the next task is only computed after the plan executor informs the motion planner that the plan has been executed. 
The motion planner state machine is defined as an RTA module. Depending on the decisions made by the associated decision module, the plan is computed by an advanced or safe planner. Computing the plan can be done by calling external functions for the safe and advanced controllers, for example, using functions from motion planning libraries such as the Open Motion Planning Library (OMPL)~\cite{ompl}. 
%

%

When a plan is computed, it is forwarded to the plan executor. The plan executor is a state machine that implements another RTA module. For each step of the plan,   the plan executor  consults a decision module on what type of controller to use. For example, in the surveillance application, if a step is leading to a collision with another robot, the plan executor will use a safe controller to guide the robot around the other robot. If the battery level is low, the safe controller might decide to first go to the charging station before going to the next point given by the plan.   
When the plan is executed, the plan executor informs the motion planner and waits for the next plan to be sent by the motion planner.

\paragraph{ROS Integration.}
The software stack is built on top of a software interface given as a library of C++ modules.  The library contains all foreign functions that implement the monitors for the decision modules and the safe and advanced controller for the RTA modules. We chose C++ for writing the external function because P programs can be compiled in to C++ programs, which in turn can compiled to ROS executables. To build the ROS executables, we used the Catkin build system (\url{http://wiki.ros.org/catkin/conceptual_overview}).  Catkin build system (popularly used for ROS projects) contains a source space where programmers include their source code. We have modified it so that this source space can support P files. This is done using the P compiler and Cmake to compile the P programs into executables that can be run on ROS.

\section{Case Studies}
\label{sec:case_studies}
\setlength{\belowcaptionskip}{-6pt}

We present two case studies with multiple runtime assurance modules. We use the case studies to show how to use SOTER on ROS framework to build safe robotics that provide safety guarantees while maintaining performance, and  to demonstrate the re-usability of the framework on a variety of robotics applications. In SOTER, application task planners, the Cpp module layer, and their RTA modules, need to be implemented independently for each application. The software stack in SOTER on ROS is largely reusable for many applications. The implementation and videos of these case studies can be found on \url{https://github.com/Drona-Org/SOTERonROS}.  

\subsection{Drone Surveillance Protocol}

\begin{figure}[t]
  \centering
  \includegraphics[scale=0.27]{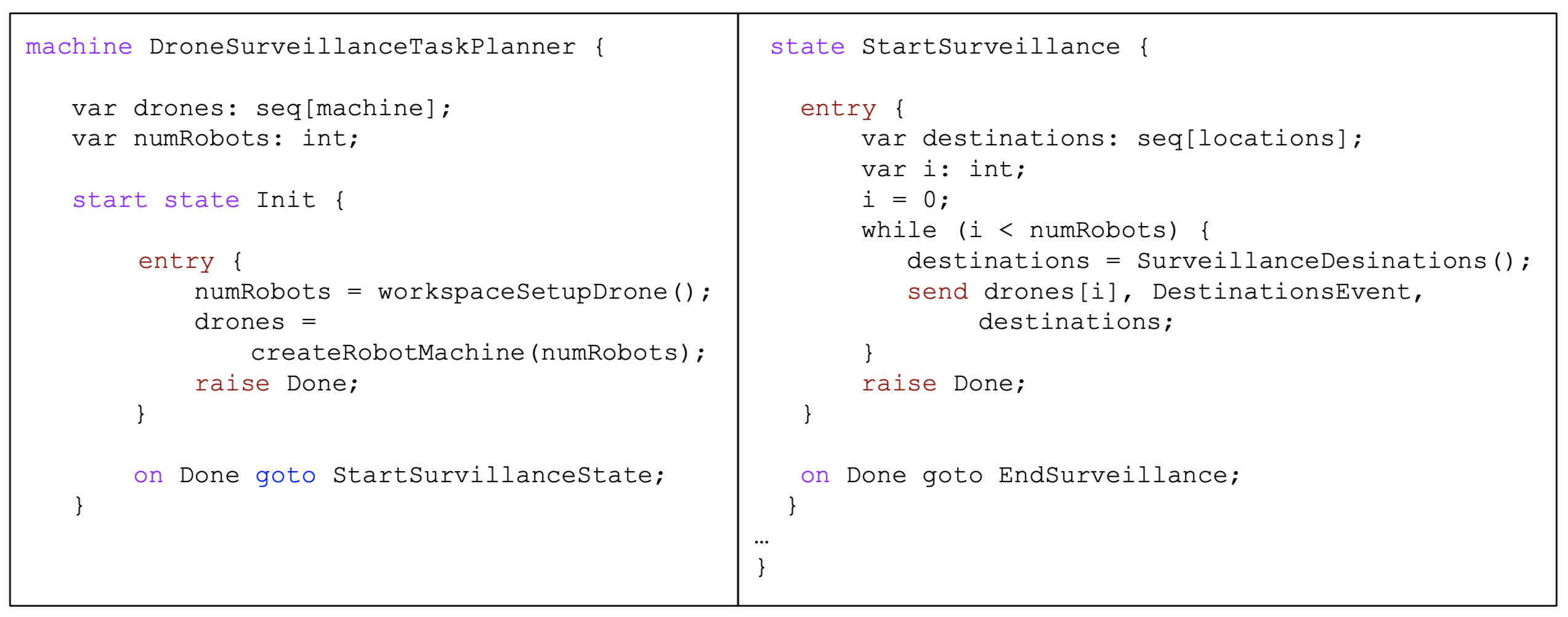}
  \caption{Application level code for Drone Surveillance Protocol. This is a simplified version of the task planner. For the full P state machine we refer the reader to the surveillance application directory on the frameworks webpage.}
    \label{fig3: Task Planner Code}
\end{figure}

The Drone Surveillance Protocol case study demonstrates how to develop a SOTER on ROS application. In this case study, there is a single drone exploring a 5x5 grid with 4 walls on the boundaries of the workspace. The goal of the drone is to explore the workspace (visit a series of locations), while ensuring the drone does not collide with the walls. To implement the case study, the programmer must first implement the application level goals in the task planner, which is implemented as a P state machine. The task planner, as depicted in  \Cref{fig3: Task Planner Code}, consists of two states, the initialization state (\texttt{Init}) and the surveillance state (\texttt{StartSurveillance}). The former is used to initialize the relevant workspace information and the robots within the application. The surveillance state is used to send destination information to the different robots, which in the case of the machine in \Cref{fig3: Task Planner Code} is done in the order of the robot's id's. Here, \texttt{DestinationsEvent} is the event queued into the robot machine, and \texttt{destinations} is the corresponding payload of the event. 

The P state machine implementing the robot machine in the drone's software stack is responsible for setting up communication with the drone's own motion planner, and initializing with ROS, which is done using a foreign function $\lstinline{RobotROSSetup()}$ that the programmer implements to connect the P machine with its ROS node. The robot machine forwards the destination point from the task planner to the motion planner, which then computes a series of way points to reach that destination. In our case studies, we use the Open Motion Planning Library's motion planner \cite{ompl}, and make it accessible in P using a foreign function. Finally, these sequence of way points are sent to the drone's plan executor, that physically executes this plan on the drone. It does so, using a series of foreign functions from the C++ modules that implement the drone's controllers, which is provided by the programmer. 

In our case study, the motion planner has no knowledge of the obstacles present in the workspace. As a result, the drone occasionally visits points that are very close to the walls. In order to ensure the drone does not collide with the walls, we construct a collision avoidance runtime assurance module. The RTA module defining the plan executor guides the robot to visit a series of locations across the workspace. The decision module monitors the location of the drone, and specifically checks to see if the next way point is located in a problematic location on the workspace. The decision module also has a parameter $\Delta$, that has the ability to look ahead to the next $\Delta$ way points of the drone's current motion plan and confirm none are in dangerous locations in the workspace. If the decision module finds that the one of the next $\Delta$ way points bring the drone too close to one of the walls, it transfers control to the safe controller. The safe controller brings the drone back to safety in the middle of the workspace. The decision module is able to perform this look ahead and return an answer in a non-substantial amount of time (near instantaneous). 

This decision module is implemented in the decision module P state machine, where the programmer implements their decision logic to determine if the robot is in a safe/unsafe state. The plan executor communicates with this decision module machine to determined whether to execute the AC or the SC. 

\begin{figure}%
    \centering
    {\includegraphics[width=4.33cm]{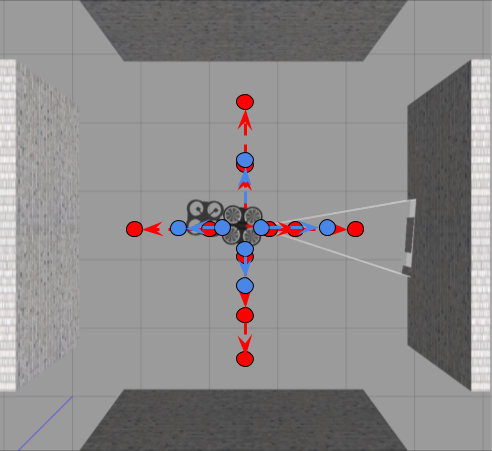} }
    {\includegraphics[width=7.7cm]{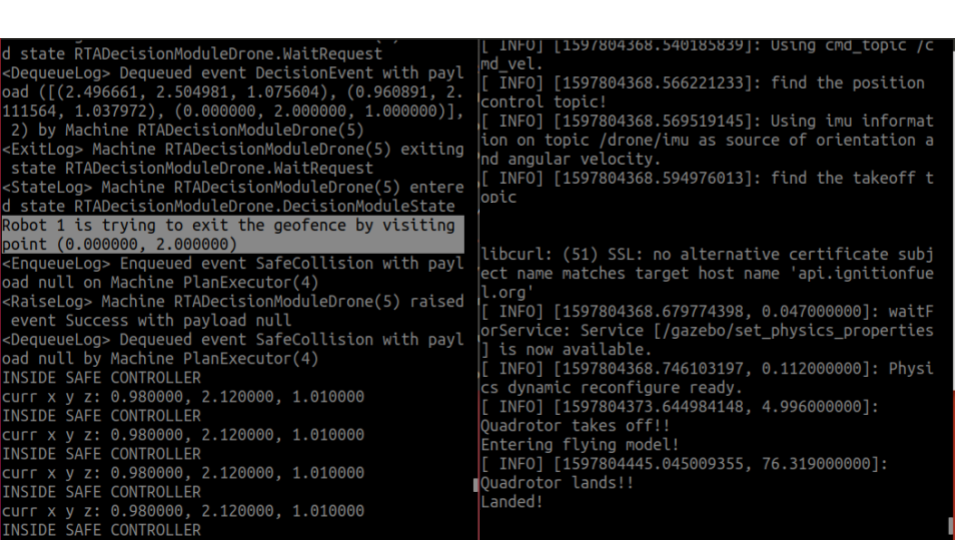} }
    \caption{Drone Surveillance Protocol.}%
    \label{fig2:Drone Surveillance Protocol}%
\end{figure}

Figure \ref{fig2:Drone Surveillance Protocol} contains a snapshot of the simulation and the terminal used to run our application. We execute our application by first launching our gazebo simulator in one window (right) and executing the ROS executable (left). The ROS executable also displays application related information such as the controller that is being used and the reason why the decision module decided to switch to one of the controllers. 
We also demonstrate the effect of the $\Delta$ parameter of the decision module in our case study. Increasing values of $\Delta$ cause the decision module to look ahead further into the drone's motion plan, and in turn makes its behavior more conservative in how close the drone can fly near the walls. Figure \ref{fig2:Drone Surveillance Protocol}, compares the drone's path with $\Delta = 1$ in red and $\Delta = 2$ in blue.

\subsection{Robot Delivery}

In the Robot Delivery case study, we demonstrate the ability to have multiple robots running asynchronously using decision modules over multiple monitors. There are two robots (TurtleBot3) that explore a grid with static obstacles. The goal of the robots is to randomly visit points on the grid indefinitely, while avoiding the static obstacles. The task planner sends each robot randomized empty destinations on the grids. Each robot has its own copy of the motion planner and plan executor. The robot machine forwards destination information to the motion planner, which in this case is the third party Open Motion Planning Library's motion planner \cite{ompl}. The motion planner computes way points to reach this destination while avoiding the static obstacles of the workspace. The motion planner also forwards the plan to the plan executor to execute. This process occurs concurrently on both robots so multiple robots can simultaneously navigate the workspace. 

In this case study, we define a decision module with 3 different runtime monitors: (1) Battery Safety, (2) Geo Fencing, and (3) Collision Avoidance. 

The first monitor is battery safety, where we prioritize safely bringing the robot to its charging station. Here our advanced controller is a node that computes control information given the current motion plan and drives the robot to the next way point in the plan. The safe controller is a certified planner that safely brings the robot to its corresponding charging station from its current position. The decision module observes battery percentage at each way point to ensure whether there is sufficient battery for executing the next $\Delta$ way points. 

The geo-fencing monitor checks whether the robot moves outside of our 5x5 grid. The RTA using this monitor can then ensure that the robot does not navigate to this region. Here our advanced controller is a node that computes control information given the current motion plan and drives the robot to the next way point in the plan. The safe controller prevents further execution of the plan and skips to the next destination, ensuring the robot remains in the safe region. The decision module observes the next $\Delta$ at each step, and determines whether the current robot would eventually explore an unsafe region. 

The third safety guarantee is collision avoidance. In the event that the two robots are simultaneously trying to visit the same destination, we ensure that a collision does not occur. The advanced controller executes the current motion plan way point by way point. The safe controller has one of the robots wait until the other finishes reaching the destination, and then proceeds. The decision module observes the next $\Delta$ way points of both robot, given their current location and motion plan, and determines whether a collision is imminent. 

In this case study, the decision module machine has 3 different aspects of the robot it must monitor simultaneously. Each RTA module also has its own AC and SC and each of the RTA modules must be composed to provide the desired security guarantees. Hence, the decision module must have an implicit prioritization of the 3 monitors, to decide which safe controller the Plan executor must run in the event multiple monitors report unsafe. In our case study, we prioritized the RTA modules in the following order: collision avoidance, geo-fencing, and battery safety. The decision module is able to perform this monitoring task and return an answer  non-substantial time.

\section{Related Work}
\label{sec:related_work}
There is a rich body of work on the design of runtime assurance components for safety-critical systems \cite{DBLP:conf/rv/DesaiDS17,DBLP:conf/aips/HofmannW06,rosrv,javamac,DBLP:conf/nfm/MassonGWCCT18,modelplex,monitoringrobotics,componentsimplex,RuntimeAF}. Some of these works present  language-based approaches that instrument an implementation of a system to assure that certain executions are enforced to satisfy certain requirements,  other approaches combine design time techniques with runtime verification techniques to assure that environment assumptions made at design time also hold at runtime \cite{DBLP:conf/rv/DesaiDS17}. 
For black-box robotic systems, or  robotic systems that include off-the-shelf machine-learning-based components that are hard to verify at design time,  Simplex-based approaches like SOTER are more suitable. Frameworks based on the Simplex (or Simplex-like) architecture include those presented in  \cite{modelplex,componentsimplex,DBLP:conf/nfm/PhanG0PSS20,RuntimeAF}. These frameworks are however designed for single component systems, or wrap the entire system using a single Simplex module, making the design of monitoring components for a distributed setting extremely difficult and complicated. In comparison,  with SOTER, we provide a framework for the design of Simplex-based RTA modules for distributed robotic systems that builds on a formally verified robotic software stack and is compatible with a variety of robot SDK's. 
 We also note that decision modules in SOTER allow for a principled and safe way to switch back from the safe controller to the advanced controller to keep performance penalties to a minimum~\cite{soter}; subsequently, an alternative approach  for realizing the reverse switching mechanism was presented by Phan et al.~\cite{DBLP:conf/nfm/PhanG0PSS20}.

\section{Outlook}
With SOTER we presented a framework for building safe distributed robotics with integrated runtime assurance modules. SOTER separates the implementation of the robot logic from the underlying robot SDK making it compatible with many robotic platforms.  For now combining multiple monitoring decisions of the decision modules is still a manual step. For the future, we plan on providing a systematic way to coordinate the different decisions. We also plan to integrate the framework with a broader class of robotics platforms, planning tools, simulators, and techniques such as introspective environment modeling~\cite{seshia-rv19}.  

\subsection*{Acknowledgments}
This work is supported in part by NSF grant CNS-1545126, the DARPA Assured Autonomy program,
Berkeley Deep Drive, and by the iCyPhy center.

\bibliographystyle{splncs04}
\bibliography{lit.bib}

\end{document}